# Image-based Automated Species Identification: Can Virtual Data Augmentation Overcome Problems of Insufficient Sampling?

Morris Klasen[1], Dirk Ahrens*[2], Jonas Eberle[2,3], and Volker Steinhage[1]

[1] *Department of Computer Science IV, University of Bonn, Endenicher Allee 19A, 53115 Bonn, Germany*

[2] *Zoologisches Forschungsmuseum Alexander Koenig, Zentrum für Taxonomie und Evolutionsforschung, Adenauerallee 160, 53113 Bonn, Germany*

[3] *Paris-Lodron-Universität, Zoologische Evolutionsbiologie, Hellbrunner Straße 34, 5020 Salzburg, Austria*

*\* Correspondence to be sent to: Zoologisches Forschungsmuseum Alexander Koenig, Zentrum für Taxonomie und Evolutionsforschung, Adenauerallee 160, 53113 Bonn, Germany; E-mail: d.ahrens@leibniz-zfmk.de*

## Abstract

Automated species identification and delimitation is challenging, particularly in rare and thus often scarcely sampled species, which do not allow sufficient discrimination of infraspecific versus interspecific variation. Typical problems arising from either low or exaggerated interspecific morphological differentiation are best met by automated methods of machine learning that learn efficient and effective species identification from training samples. However, limited infraspecific sampling remains a key challenge also in machine learning.

In this study, we assessed whether a two-level data augmentation approach may help to overcome the problem of scarce training data in automated visual species identification. The first level of visual data augmentation applies classic approaches of data augmentation and generation of faked images using a GAN approach. Descriptive feature vectors are derived from bottleneck features of a VGG-16 convolutional neural network (CNN) that are then stepwise reduced in dimensionality using Global Average Pooling and PCA to prevent overfitting. The second level of data augmentation employs synthetic additional sampling in feature space by an oversampling algorithm in vector space (SMOTE). Applied on two challenging datasets of scarab beetles (Coleoptera), our augmentation approach outperformed a non-augmented deep learning baseline approach as well as a traditional 2D morphometric approach (Procrustes analysis).

**Introduction**

Correct species identification and delimitation is fundamental for taxonomy but also for many other biological disciplines including medicine and food production. Robust estimation of species limits defined by intra- versus interspecific variation requires adequate sampling of both levels to recognize these boundaries. This includes quantitative but also qualitative approaches.

However, in biodiversity studies of taxonomically poorly known groups and ecosystems (e.g., Pons et al. 2006; Monaghan et al. 2009; Lim et al. 2012; Fujisawa and Barraclough 2013; Tang et al. 2014; Ahrens et al. 2016), sampling may vary due to actual differences in species abundances or due to incomplete sampling. Many species are rare and may be represented by only one "singleton" exemplar (McGill et al. 2007).

Species only known from a single specimen are common in biodiversity samples. Reviews suggest that in tropical arthropod samples, 30% of all species are represented by only one specimen (Bickel 1999; Novotny and Basset 2000; Coddington et al. 2009), and additional sampling did help only little eliminating such rarity (Scharff et al. 2003; Coddington et al. 2009).

Ecologists have long been aware and fascinated by the phenomenon of rarity in biodiversity (Magurran and Henderson 2003; Scharff et al. 2003; Cunningham and Lindenmayer 2005; Mao and Colwell 2005; Chao et al. 2009) which is not surprising given that rare species are particularly important from the points of view of conservation, ecology, and evolutionary biology. Such species are frequently also the focus for policy makers (Prendergast et al. 1993; Soltis and Gitzendanner 1999).

Thus, this matters principally in all species recognition approaches, such as morphometrics or DNA-based species delimitation (Lim et al. 2012; Ahrens et al. 2016) but expectedly also in recently employed automated image recognition systems for taxon identification (Van Horn et al. 2018). Given the particular data quality (digital image data, DNA barcodes, etc.) the effect of poor sampling and its proportion may be even higher in those studies compared to morphology-based alpha taxonomy studies, where usually all known specimens may be included.

The issue is so challenging not only because the diversity of over 1 million recorded insect species is immense, but also because in some cases, the human eye cannot distinguish between closely related species at all.

Over the years, many systems have been introduced, attempting to automate visual identification of rare and numerous species (e.g., Feng et al. 2016; Yang et al. 2015; Steinhage et al. 2007; Watson et al. 2004). However, small sample sizes of individual classes are typically problematic in deep learning (Chawla et al. 2004). There are concerns with validation, metrics and overfitting (i.e., the trained system has learned the samples too well due to the high number of descriptive features, and therefore will not generalize sufficiently on new unseen samples). There have been approaches analyzing and solving the issue of small sample sizes: Perez and Wang (2017) investigated several augmentation strategies with conventional methods as well as Generative Adversarial Networks (GAN) in order to improve performance on small datasets. Schonfeld et al. (2019) and Xian et al. (2019) used zero/few-shot-learning which trains the network only on species with (sufficient) samples to reduce the risk of overfit and aim to recognize unseen species. Recent advances in computer vision and deep learning lead to more automated systems for image-based classification, e.g. transfer learning (Zheng et al. 2016) or feature extraction for fine-grained classification (Zhang et al. 2017). While Convolutional Neural Networks (CNN) achieved tremendous progress in image classification tasks, training the ever increasing network sizes becomes a computational and cost constraint. Razavian et al. (2014) and Valan et al. (2019), however, have shown that pretrained networks can be employed to achieve similar performance to self-trained models. In this study, we will apply transfer learning in both ways (1) *feature transfer*, i.e., using the pretrained weights of the model as it is, or (2) investing extra computation by *fine-tuning* the network weights on the given specific dataset to optimize the identification performance. Additionally, we will apply and evaluate the two-step data augmentation in terms of (1) classical augmentation methods and faked images generated by Generative Adversarial Networks (GAN) (Goodfellow et al. 2014) and (2) synthetic oversampling in feature space (SMOTE). Alongside with the data augmentation we employ several methods of reducing the number of describing features, i.e., global pooling and principal component analysis, because a training of a species identification system given few data samples in high-dimensional descriptive features spaces will result in overfitting.

We will evaluate the data augmentation methods by comparison with (1) traditional studies applying a 2D geometric morphometric distinction between samples based on hand-digitized semilandmarks and with (2) a deep learning baseline species identification approach based on bottleneck-features generated by a deep convolutional network not using the aforementioned augmentation methods.

**Material and methods**

*Datasets*

The study uses two empirical image datasets of scarab beetle genitalia (Coleoptera: Scarabaeidae) that were previously investigated with traditional geometric morphometric approaches (Özgül-Siemund and Ahrens 2015; Eberle et al. 2015). They exhibit challenging conditions of differentiation and identification, including very low number of samples per species and imbalanced sample distribution. Genital organs are commonly used as standard diagnostic features in insect taxonomy. The first dataset (image size: 250x250 pixels) of 112 specimens and 7 species includes images of little differentiated female genitalia of the South African genus *Pleophylla* (Özgül-Siemund and Ahrens 2015; Eberle et al. 2017). Genitalia were displayed with a centered view and homogeneous viewing-angle (90°) and magnification. Central object of each image is the distal sclerite at the base of the bursa copulatrix of the female genitalia (Özgül-Siemund and Ahrens 2015), which is surrounded by membranous tissue. Many specimens (Fig. 1) featured sclerotized parts (of the same color and texture as the distal sclerite) in the lateral image regions, which, in ca. 5% of the cases partially occlude the distal sclerite.

The second image dataset (image size: 600x400 pixels) of male copulation organs of the African genus *Schizonycha* was composed of 76 specimens from 33 species (Eberle et al. 2015). Again, all images were centered, oriented in the same direction and filling out most of the image area without any disturbing background. The dataset contained 13 species with one specimen and 14 species represented by 2 specimens each. Here, we excluded the 13 one-sample species because we cannot validate our methods on them using a train/test split.

The original studies (Eberle et al. 2015; Özgül-Siemund and Ahrens 2015) measured overall 2D geometric morphometric distinction between samples based on hand-digitized semilandmarks that were here re-analyzed using Procrustes analysis (Gower 1975; Rohlf and Slice 1990).

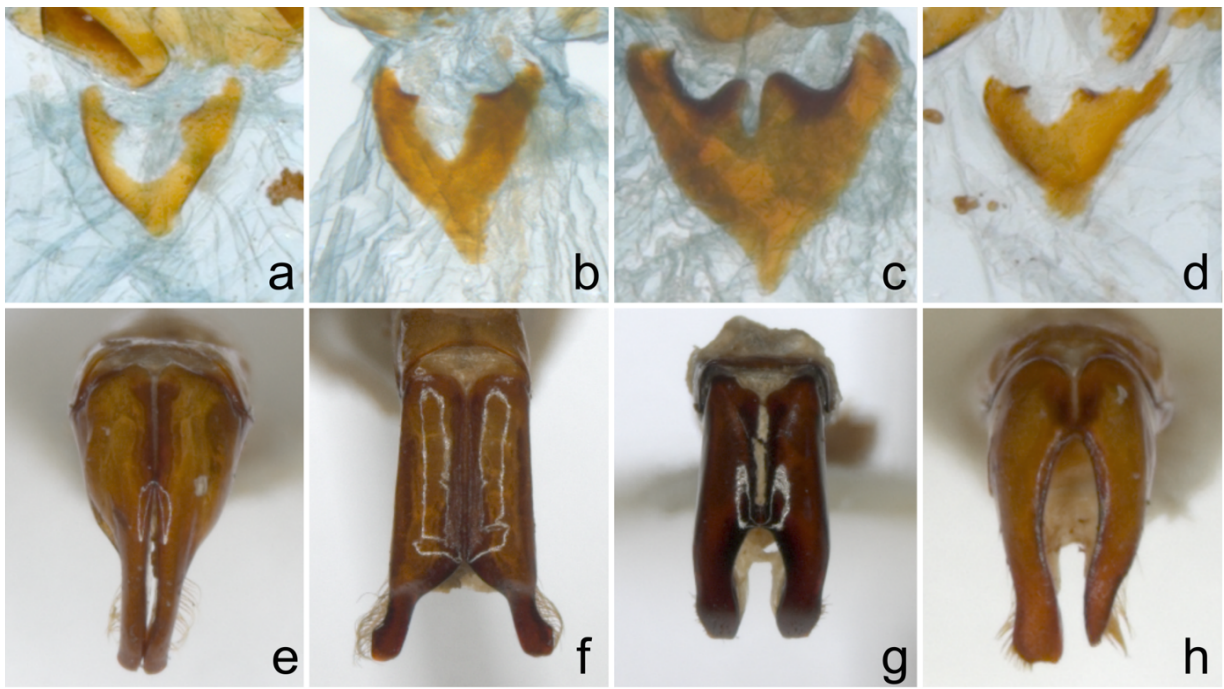

**Figure 1.** Examples of specimen images: a- *Pleophylla pilosa*, b- *P. ferruginea*, c- *P. navicularis*, d- *P. warnockae*, e- *Schizonycha gracilis*, f- *S. neglecta*, g- *S. durbana*, h- *S. transvaalica*.

*Workflow for automated species identification*

The overall workflow (Fig. 2) had two endeavors that must work hand in hand to perform species identification given small sizes of specimen samples: (1) steps of increasing the number of samples by augmentation, fake generation and oversampling; (2) steps of reducing the number of describing features by global pooling and principal component analysis to avoid overfitting.

The input of the workflow were original specimen images as depicted in Fig. 1 and had the following steps. First, two augmentation steps were applied to these original specimen images: (1) classic data augmentation methods and (2) generation of new fake samples using Generative Adversarial Networks (GAN). The original image samples as well as the newly generated image samples by augmentation and GAN (Table 1) from the input of pretrained VGG-16 (Simonyan and Zisserman 2014) convolutional neural network layers to generate descriptive feature vectors. These high-dimensional feature vectors are called bottleneck features. To avoid overfitting, global average pooling reduces dimensionality of the bottleneck features significantly up to a factor of 50 (see below). The next step implemented a fine-tuning of the derivation of the VGG-16 bottleneck features by presenting training data of the observed species of the genus *Pleophylla* and the genus *Schizonycha*. Technically, we connected the bottleneck feature neurons to a fully connected layer with $S$ neurons - where $S$ is the number of the observed species. This allows the fine tuning by presenting this network all training data of the genus *Pleophylla* and the genus *Schizonycha* consisting each of an input image and the corresponding species label. To avoid overfitting we applied then Principal Component Analysis (PCA) reducing the 512 descriptive features to a smaller number of descriptive features with a remaining cumulative trait variation (CTV) compared to the original feature vectors. In this vector space of reduced dimensionality a synthetic oversampling algorithm (SMOTE) (Chawla et al. 2002) was applied to generate new samples. But SMOTE generated new samples in terms of feature vectors and not in terms of image samples. Finally, a linear support vector machine (SVM) (Suykens and Vandewalle 1999) was trained on all resulting feature vectors of reduced dimensionality for species identification. In the following we describe explicitly all components of the workflow.

**Table 1.** Characteristics of the original and augmented data (number of samples) of the two study cases of *Pleophylla* and *Schizonycha*.

|  | Original | Rotation | StyleGAN | SMOTE |
|---|---|---|---|---|
| ***Schizonycha*** | | | | |
| Range | 2-7 | 16-42 | 20 | 0-36 |
| Average | 3 | 23 | 20 | 21 |
| Total | 62 | 496 | 420 | 449 |
| ***Pleophylla*** | | | | |
| Range | 2-41 | 8-160 | 20 | 1-20 |
| Average | 16 | 61 | 20 | 8 |
| Total | 112 | 432 | 140 | 54 |

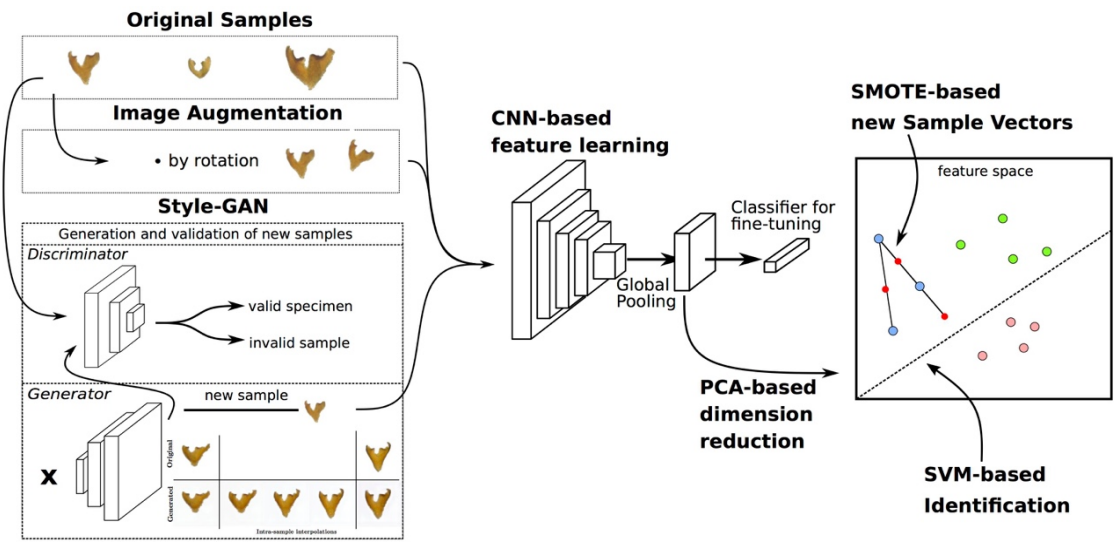

**Figure 2.** Overview of the overall workflow of automated species identification and image augmentation from sample to ID.

*Image data augmentation*

Since size and (a)symmetry represent crucial parameters of genitalia morphology, we refrained augmentation procedures like zooming, affine transformations and mirroring. Thus, image rotation was the only valid transformation. Overfitting that may occur in small data sets was prevented by sparsely rotating images by 20° with 5° intermediate steps.

*Generative Adversarial Networks for generating new interpolated samples*

Generative adversarial networks (GAN) have been introduced by Goodfellow et al. (2014). GANs create new data samples - so-called faked samples - that resemble given training data. Most prominent applications are for example, faked photographs of human faces that do not belong to real persons (Karras et al. 2018), fake paintings (Elgammal et al. 2017), fake lead sheet music (Liu and Yang 2018) etc. GANs have two components: the generator generates fake samples that become negative training examples for the second component, the discriminator. The discriminator learns to distinguish the generator's fake samples from real data samples and penalizes the generator for producing implausible results. By incorporating this feedback from the discriminator, the generator in turn learns to improve the generation of fake data to fool the discriminator by classifying the generator's fake output as real.

In our specific case, we applied StyleGAN (Karras et al. 2019), an adaptation of the classic GAN approach building upon the Progressive-GAN (Karras et al. 2018) with a latent input mapping network and intermediate layer inputs for more control authority over generated image styles and better quality.

*Convolutional Neural Network for generating descriptive features*

Convolutional Neural Networks (CNN) are having great success recently in large-scale image recognition (Krizhevsky et al. 2012; Zeiler and Fergus 2013; Sermanet et al. 2014). We chose the VGG-16 (Simonyan and Zisserman 2014) as a network approach, regularly used for feature extraction because of its high-level representation learning with a generic and linear network structure. We employed the pretrained VGG-16 network whose parameters are derived in the ImageNet competition (Russakovsky et al. 2015), one of which the VGG-16 won in 2014, trained on over 1.3 million images across 1000 classes. The resulting bottleneck features are constituted as 512 feature matrices of size $7 \times 7$ each. Linearization of these features would result in 25088-dimensional feature vectors. Feature vectors of more than 20.000 dimensions would be hugely overdetermined - especially for small dataset applications. Therefore, we applied global average pooling (Lin et al. 2013) on the 512 feature matrices of size $7 \times 7$ returning the average values of all $7 \times 7$ feature matrices yielding 512-dimensional feature vectors (and therefore a dimensionality reduction of a factor 49) where each feature vector corresponds to an original image samples or to an image sample generated by rotational or fake augmentation. Up to that point, the VGG-16 network has no information about the given task of species information. This happens in the fine-tuning of the VGG-16 network parameters by adding a fully connected layer of *S* neurons where each of the *S* neurons corresponds to one species and is connected to each of the 512 feature neurons derived by the global average pooling. Training is now performed solely for the fully connected layer in 50 epoches to adjust the parameters of the connections between the 512 feature neurons and the N species neurons. Then all parameters of the complete VGG-16 are fine-tuned using adaptive learning rate optimization (Kingma & Ba, 2014) with a learning rate of 1e-5.

*PCA for dimensionality reduction*

Finally, we applied principal component analysis (PCA) (Abdi and Williams 2010) to reduce dimensionality that transforms the 512 potentially correlated features into a smaller number $N$ of uncorrelated features (principal components) that are the eigenvectors of the features' covariance matrix. Instead of empirically deciding the optimal number $N$ of features, we used maximum accuracy along cumulative trait variation (CTV) of the resulting principal components as criterion to estimate the feature space (PC axes) employed in species identification. More specifically, the CTV was evaluated in steps of 10% with respect to the identification accuracy (using Support Vector Machine, see below) to infer the optimal remaining cumulative trait variation to be used in subsequent species identification analysis (Fig. 4). This way the amount of retained information (i.e., number of principal components) was dataset-specific to optimize the identification accuracy values (see Table 2).

**Table 2.** Comprehensive overview of accuracy of the out-of-the-box results with a SVM for classification of CNN image analysis compared to traditional 2D morphometric data (Procrustes analysis). Improvements ($\Delta_{x-0}$) through data augmentation (rotation/ Style-GAN/ SMOTE) for each single and combined approaches compared to the original data (VGG-16 pretrained) along the preferable amount of CTV used (%) are shown.

|  | *Pleophylla* | | | *Schizonycha* | | |
| --- | --- | --- | --- | --- | --- | --- |
| **Method** | Accuracy (%) | $\Delta_{x-0}$ | CTV (%) | Accuracy (%) | $\Delta_{x-0}$ | CTV (%) |
| **CNN** | | | | | | |
| **VGG-16 pretrained** | 65.15 | NA | 90 | 46.77 | NA | 90 |
| **VGG-16 fine tuned (VGG-ft)** | 68.07 | 2.92 | 90 | 57.41 | 10,64 | 90 |
| **VGG-ft + Rotation** | 74.22 | 6.14 | 90 | 62.90 | 5,49 | 90 |
| **VGG-ft + Style-GAN** | 74.83 | 6.75 | 90 | 80.32 | 22,91 | 90 |
| **VGG-ft + Rot + GAN+SMOTE** | **80.07** | **12.0** | 90 | **85.16** | **38.39** | 90 |
| **Procrustes** | | 74.64 | NA | 70 | 74.51 | NA | 90 |

*Oversampling for rebalancing the dataset*

The image samples are imbalanced between taxa with many specimens and those with few specimens (Table 1). Therefore, we used the Synthetic Minority Oversampling Technique (SMOTE) algorithm (Chawla et al. 2002) to oversample minority data (i.e. little represented species) in the *N*-dimensional feature vector space resulting from PCA. SMOTE oversamples minority data by generating minority-class interpolations between existing samples in the *N*-dimensional feature vector space. The SMOTE oversampling (1) re-balances the class-distribution to strengthen underrepresented species; (2) inflates the dataset further by creating quasi interpolations of existing samples in the vector space; (3) supports the SVM based identification (see below) by decreasing the imbalance of the training data with respect to the number of specimen per species.

*Support Vector Machine for species identification*

As a classifier for our descriptive *N*-dimensional feature vectors (derived by the PCA) we used a Support Vector Machine (SVM) (Suykens and Vandewalle 1999). It is typically chosen in applications with small datasets and high dimensionality of the data because of its overfitting resistance and robustness with respect to imbalanced datasets (He and Garcia 2009). The SVM does so by finding linear decision boundaries in the *N*-dimensional vector space to separate the classes. Because of the nature of our small datasets, with very few specimen for some species, we chose, instead of using the regular hinge-loss for hyperplane optimization, to use squared-hinge loss SVM

$$\sum_{i=0}^{n} max\ (0.1 - y_i(wx_i + b_i))^2$$

to maximize misclassification penalties. For the same reason we will also use the least squares regularization of the SVM weights $\| w \|_2^2$ instead of least absolute deviation.

$$\min \frac{1}{2} \| w \|_2^2 + C \sum_{i}^{N} max(0.1 - y_i\ (wx_i + b_i))^2$$

summarizes the whole SVM classification, with *x* being an *N*-dimensional feature vector representing the image of a specimen, $y_i$ being the supplied predetermined species label of that specimen. The remaining variables are SVM specific parameters, i.e. *w* defines the orientation of the separating boundaries between species in the *N*-dimensional vector space, $b_i$ is the species specific bias and C being the penalty parameter that can be chosen empirically.

*Relevance ranking of morphological features by heatmaps*

Highly similar beetle or insect species are used to be distinguished by the shape of their genitalia, in males and females (Eberhard 1985; e.g., Özgül-Siemund and Ahrens 2015). Traditionally this is done by trained experts, who extrapolate gaps in shape variation to assume species boundaries, and interpolate from related taxa for rare species for which only a few or a single specimen are available. Crucial for this purpose is for them to find the important diagnostic features.

Class activation maps (CAM) (Zhou et al. 2018) offer a simple way to visualize the discriminative image regions used by CNN-based approaches to identify object classes in images. Given the VGG-16 convolutional layers in our workflow, CAM visualizes the activations of the bottleneck features of a particular classified object in terms of a heatmap. The heatmap has therefore the size of the feature maps of the bottleneck features, i.e. in VGG-16 feature maps of size 7 × 7. To impose the heatmap on the input image, the heatmap is upscaled to the size of the input image and then overlay it on top of the input (cf. Fig. 3). These show, overlayed on top of the specific specimen with an intensity signature, which image regions in particular the network used to predict the species. These underlying structures of interest in the depicted specimen could be homogenous pixel regions, edges or combinations with varying colors, sizes, locations and orientation. The color intensity of the heatmap implies a relevance ranking, where dark red regions are more important (i.e., informative) than orange, green, or even blue ones. Warm colors thus indicate diagnostic traits of the specimen.

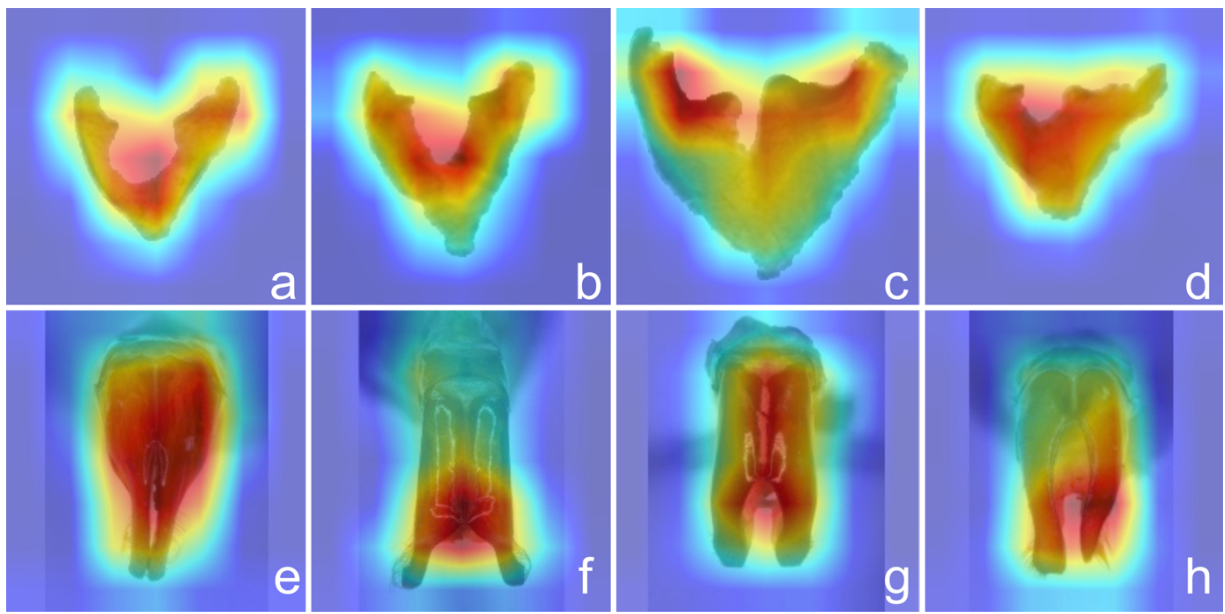

**Figure 3.** Heatmaps of specimen images: a- *Pleophylla pilosa*, b- *P. ferruginea*, c- *P. navicularis*, d- *P. warnockae*, e- *Schizonycha gracilis*, f- *S. neglecta*, g- *S. durbana*, h- *S. transvaalica*.

*Integration of methods in the workflow and overall evaluation*

The order of steps, i.e. the method applications, in the workflow is highly relevant. When augmenting the number of actual images and adding StyleGAN interpolations, we have to concatenate them with the original dataset before deploying the 512-dimensional feature vectors derived as CNN-Codes plus global average pooling. PCA must be applied on the 512-dimensional vector space yielding an *N*-dimensional feature space ($N<512$). SMOTE oversampling is applied in this *N*-dimensional feature space.

Due to the very limited amount of samples per class (i.e., specimen per species), a simple test/validation split of the dataset was not adequate to evaluate the following experimental results. Therefore the chosen evaluation method was 10 times repeated stratified k-fold cross-validation with *k = 2*. We calculate the identification accuracy by comparing the predicted species class to the actual species class:

$$(identification) accuracy = \frac{TP + TN}{TP + TN + FN + FN}$$

with TP = number of true positive identifications, FP = number of false positive identifications, TN = number of true negative identifications, FN = number of false negatives identifications.

**Results**

*Data augmentation and accuracy*

The proposed approach of generating descriptive features for species identification by utilizing augmented data derived by rotational variations, GAN- and SMOTE-generated data revealed clear improvements with respect to the identification accuracy compared to the standard feature generation by a pretrained and fine-tuned VGG-16 just based on the original data. Additionally, the proposed approach outperformed a species identification using 2D geometric morphometric features from hand-digitized semilandmarks using Procrustes analysis (Table 2).

We obtained optimal results (i.e., with maximum accuracy) for pretrained VGG-16 data with 90% CTV in *Pleophylla* and near optimal results for *Schizonycha* (Fig. 4). The latter had, however, its highest accuracy already at 40% CTV. When data were fully augmented (rotation, GAN, SMOTE), the two datasets had similar relationships between identification accuracy and CTV, showing optimal identification accuracy with 90% CTV. For morphometric data, maximum accuracy was found at 70% and 90% CTVs in *Pleophylla* and *Schizonycha*, respectively, and a rapid decline in accuracy was found in both datasets towards 100% CTV.

The identification accuracy just using the pretrained VGG-16 for feature generation, was rather low for both datasets (Table 2), with 65.15% and 46.77% for *Pleophylla* and *Schizonycha*, respectively. Fine-tuning of VGG-16 showed a small increase to 68.07% of identification accuracy for *Pleophylla,* while for *Schizonycha* fine-tuning showed an increase of accuracy of 10.64% compared to the pretrained results.

Augmentation and GAN-interpolated images increased performance by introducing new samples: augmenting data by adding rotated images and by increasing the class distribution imbalance improved the classification accuracy by 6.14% for *Pleophylla* and 5.49% for *Schizonycha*. For the Style-GAN image generation and regularization strategy we trained the network with settings that maximize sample diversity. To further diversify fake images and to optimize for intra-sample interpolations, we used a trigonometric mapping of our latent input vector. This way we added 20 GAN-generated images to each class, which increased classification accuracy in the *Pleophylla* dataset by another nearly 6.75%, while in the *Schizonycha* dataset Style-GAN increased classification accuracy substantially by 22.91% due to the significant increase in dataset size.

Synthetic oversampling after data augmentation using image augmentation and Style-GAN, improved identification accuracy significantly to a total of 80.07% and 85.16% for *Pleophylla* and *Schizonycha,* respectively (Table 2); we obtained an accuracy improvement of accuracy of 12.0% and 38.39% for *Pleophylla* and *Schizonycha,* respectively. This also outperformed the species identification using morphometric data of the same morphological structures (by 5.43% and 10.65%, respectively) since the accuracy of Procrustes identification yielded 74.64% and 74.51% for *Pleophylla* and *Schizonycha,* respectively (Table 2).

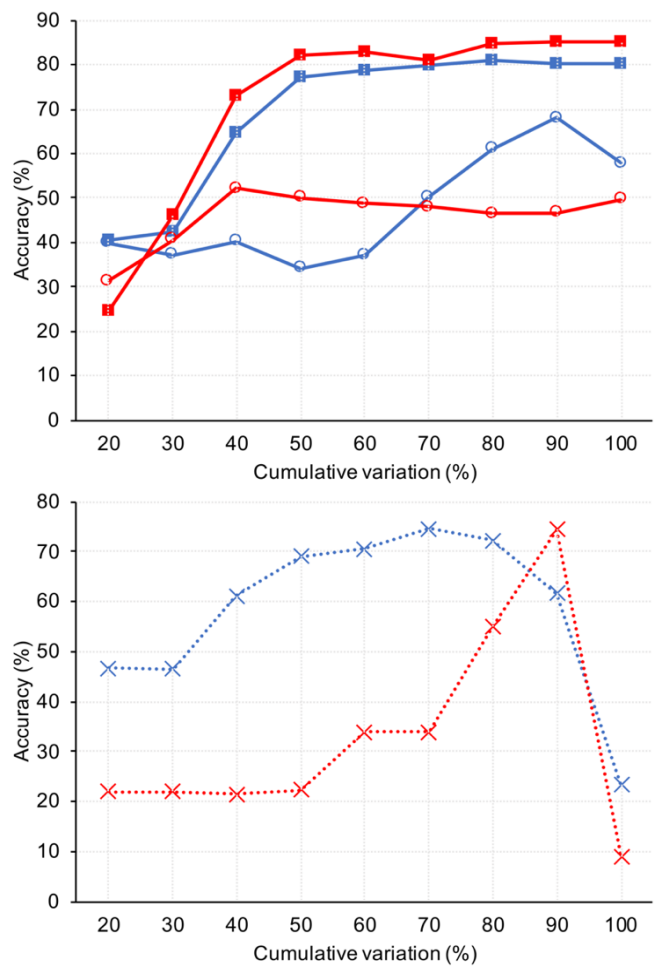

**Figure 4.** Plot of the cumulative trait variation (CTV) against accuracy for the both datasets (*Pleophylla* - blue; *Schizonycha*- red), using baseline (empty circle) and augmented data (all, incl. GAN and SMOTE; filled square) (above), compared to that derived from morphometric data (below; Procrustes analysis, empty triangle).

*Species distribution in feature space*

Two-dimensional plots of the outcome of CNN produced similar clusters compared to those of morphometric analysis using Eigenshape (Fig. 5). Actually, the information of useful variation of the CNN approaches is distributed over many more dimensions compared to hand-selected morphometric data. For the *Pleophylla* dataset, Procrustes analysis returned 25 principal components (PCs) axes with a CTV of 90% and nine PCs at 70%. In *Schizonycha*, a total of six PCs show 90% CTV (Supplement Table 1). Similarly, the here proposed approach of transfer learning, automated feature extraction and classification, resulted in 23 EV that represented 90% of total trait variation for the *Pleophylla* dataset and in 41 EV that represented 90% of total trait variation for *Schizonycha* when used for the feature generation with the pretrained VGG-16. In *Schizonycha*, morphospace (i.e., the surface of the convex hulls around the dots of a single species; Fig. 5) of each species resulted to be more extended in the CNN analysis (Fig. 5e) compared to the geometric morphometric analysis (Fig. 5d). In *Pleophylla*, no clear difference was evident (Fig. 5). In contrast to that, when data were augmented (Fig. 5c, f), the area covered by each species was even more expanded resulting in also more overlap between the species.

*Heatmaps*

As expected from directly non-homologous morphological features, i.e. male vs. female genitalia of two different genera, the heatmaps (Fig. 3) showed different hotspot zones for both datasets. In *Schizonycha*, hotspot zones comprised in all cases the part of the parameres (apex) which is actually intromitted into the female vagina, its shape being thus very likely under high selective pressure. In contrast to that, the basal portion of the phallobase was often less "informative" and only in some cases resulted as a hotspot in the heatmap (e.g., Fig. 3e, g). However, in some cases hotspots comprise also the more basal portions of parameres, in others not (e.g., Fig. 3e vs. 3f). In *Pleophylla*, the hotspots mainly resulted in the central sinuation of the sclerite (Fig. 3a), and/ or its basal projection (e.g., Fig. 3c, d).

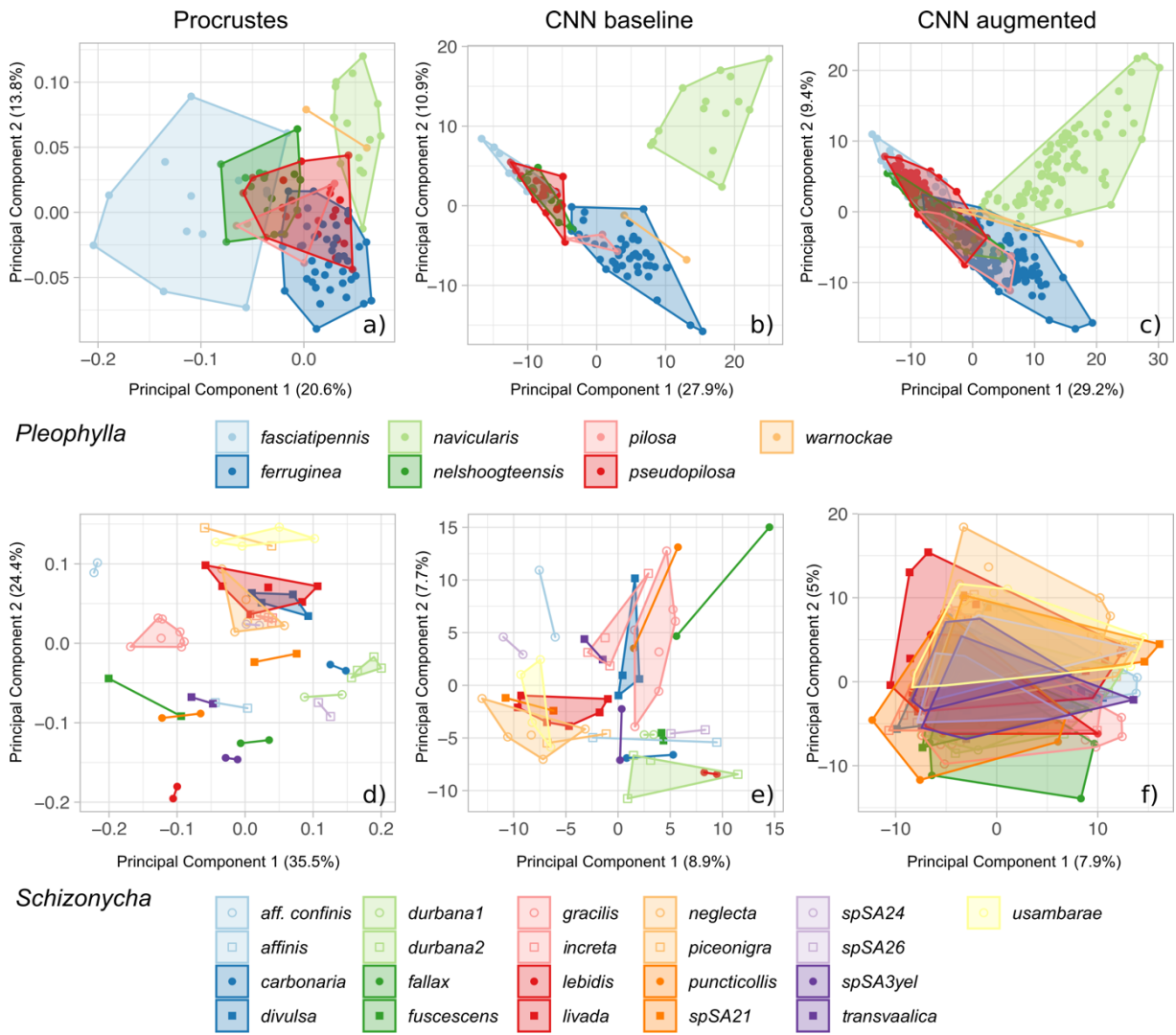

**Figure 5.** 2D scatterplots (of the most informative axes) of the outcome from CNN (PCA; baseline data, augmented data) compared to morphometric analysis based on semilandmarks in the two study cases (*Pleophylla*: a-c, *Schizonycha*: d-f).

**Discussion and Conclusions**

Our study showed for the first time the discriminative benefits of automated data augmentation used in species identification compared to men-defined traits, such as using morphometric data, as implemented in statistical approaches (Procrustes analysis). For unmanipulated baseline data (pretrained VGG-16 bottleneck features), hand-defined morphometric data were still slightly superior for the datasets used here, however, this comes at high costs (human work time for trait digitization). We showed that our proposed data augmentation strategy considerably increased classification accuracy (Table 2) compared to the baseline pretrained VGG-16 feature extraction. We tested several ways to augment the sample count of weakly represented classes (species) to firmly define class-specific vector subspaces while simultaneously using the CTV to control our overfit very carefully. Thereby, it was a main criterion that faked data augmentation was within the limits of the natural variation of the respective species. These findings might be highly relevant for future application of automated species recognition in real life, particularly when being applied in areas known as biodiversity hotspots (e.g., Myers et al. 2000), where rare species are very common (Lim et al. 2012).

However, in future, more datasets should be explored in a similar way to further refine the proposed workflow, methods and conclusions across more datasets seeking to find ways to further improve the results with artificially increased sampling. The literature record on DNA taxonomy and DNA-based species delimitation shows that a wide range of empirical studies and computer simulations (e.g., Ahrens et al. 2016, Sukumaran and Knowles 2017) is crucial to fully understand pitfalls (e.g., Carstens et al. 2013) and behavior of computational algorithms. Given the commonness of rare species, it is crucial to address sampling issues in the field of deep learning and artificial intelligence for successful application of those methods in taxonomy and environmental surveying. Also investigating the impact of trait- and specimen diversity within datasets might be of high interest. Here we were confined to a small selection of classic data augmentation methods (i.e. only rotational augmentations) because of dataset specific limitations (mentioned above). The performed heatmap analysis based on CNN results points the way ahead to an automated diagnostic trait extraction, which can be visualized (La Salle et al. 2009) and used in integrative taxonomy approaches (Yeates et al. 2011).

Last but not least, analysis of a wider representation of geographical populations would be needed. Similar as with DNA data (Lohse 2009; Bergsten et al. 2012; Chambers and Hillis 2020) it is expected that geographically structured data from many populations will generate a more extended morphospace that will cover morphological variation of the species more realistically. Such extended geographical sampling is expected to be beneficial also for GAN and SMOTE data augmentations.

All in all, this study is a contribution to meet the challenge of classifying the imbalanced distribution of sample sizes across biodiversity, i.e., some species are more abundant and easier to collect than others. As Van Horn et al. (2017), the developers of iNaturalist, one of the world's most popular identification approaches for plants and animals, pointed out: "Specifically, we observe poor results for classes with small numbers of training examples suggesting more attention is needed in low-shot learning." Therefore, a major task in order to automatize and thereby speed up biodiversity research and monitoring by the application of artificial intelligence, is to continue and to boost the development of specimen collections instead of slowing it down (Prathapan et al. 2018) with the aim to overcome the shortness of specimen sampling that is inherent to the nature of many species (Lim et al. 2012).

**Supplementary Material**

Data available from the Dryad Digital Repository: https://doi.org/10.5061/dryad.r4xgxd29t

**Author contributions**

M.K., D.A., and V.S. conceived the idea and designed the study; D.A. compiled data sets; J.E. performed morphometric analyses, M.K., wrote the scripts, performed analyses, interpreted results, and prepared the first draft. All authors revised and commented drafts at different stages and contributed to the final version of the manuscript.

**Disclosure statement**

Herewith all authors confirm that there are not any conflicts of interest to disclose.

**Acknowledgements**

We are thankful to the helpful comments of the referees, as well as to the colleagues of the Museum A. Koenig who with their strategic discussions about *how to make taxonomy more applicable* helped us to focalize the idea of this study.

**Figure descriptions:**

**Figure 1.** Examples of specimen images: a- *Pleophylla pilosa*, b- *P. ferruginea*, c- *P. navicularis*, d- *P. warnockae*, e- *Schizonycha gracilis*, f- *S. neglecta*, g- *S. durbana*, h- *S. transvaalica*.

**Figure 2.** Overview of the overall workflow of automated species identification and image augmentation from sample to ID.

**Figure 3.** Heatmaps of selected specimen images that illustrate decissive areas for species identification: a- *Pleophylla pilosa*, b- *P. ferruginea*, c- *P. navicularis*, d- *P. warnockae*, e- *Schizonycha gracilis*, f- *S. neglecta*, g- *S. durbana*, h- *S. transvaalica*.

**Figure 4.** Plot of the cumulative trait variation (CTV) against accuracy for the both datasets (Pleophylla - blue; Schizonycha- red), using baseline (empty circle) and augmented data (all, incl. GAN and SMOTE; filled square) (above), compared to that derived from morphometric data (below; Procrustes analysis, empty triangle).

**Figure 5.** Scatterplots of the two most informative axes from PCAs on CNNs (original data and fully augmented data) and morphometric Procrustes analysis in both study cases (*Pleophylla*: a-c, *Schizonycha*: d-f). Percentages indicate the fraction of displayed total variation.